\pdfoutput=1

\documentclass[11pt]{article}

\usepackage[final]{acl}

\usepackage{amsmath} 
\usepackage{times}
\usepackage{latexsym}
\usepackage{url}
\usepackage{multirow}
\usepackage{enumitem}
\usepackage{adjustbox}
\usepackage{tcolorbox}
\usepackage{booktabs} 
\usepackage{amsfonts}
\usepackage[T1]{fontenc}
\usepackage[utf8]{inputenc}
\usepackage{microtype}
\usepackage{inconsolata}
\usepackage{graphicx}
\usepackage{xspace}

\setlist{leftmargin=*,nosep}

\title{Segment First, Retrieve Better: Realistic Legal Search via Rhetorical Role-Based Queries}

\author{Shubham Kumar Nigam$^{1}$ \qquad Tanmay Dubey$^{1}$ \\ 
\textbf{Noel Shallum}$^{3}$ 
\qquad \textbf{Arnab Bhattacharya}$^{1}$\\
$^{1}$ IIT Kanpur, India \quad
$^{2}$  IISER Kolkata, India \quad
$^{3}$ Symbiosis Law School Pune, India \\
\texttt{\{sknigam, tanmay, arnabb\}@cse.iitk.ac.in} \\ \quad \texttt{noelshallum@gmail.com}
}

\date{}

\begin{document}
\maketitle

\begin{abstract}
Legal precedent retrieval is a cornerstone of the common law system, governed by the principle of stare decisis, which demands consistency in judicial decisions. However, the growing complexity and volume of legal documents challenge traditional retrieval methods. TraceRetriever mirrors real-world legal search by operating with limited case information, extracting only rhetorically significant segments instead of requiring complete documents. Our pipeline integrates BM25, Vector Database, and Cross-Encoder models, combining initial results through Reciprocal Rank Fusion before final re-ranking. Rhetorical annotations are generated using a Hierarchical BiLSTM CRF classifier trained on Indian judgments. Evaluated on IL-PCR and COLIEE 2025 datasets, TraceRetriever addresses growing document volume challenges while aligning with practical search constraints, reliable and scalable foundation for precedent retrieval enhancing legal research when only partial case knowledge is available.
\end{abstract}

\section{Introduction}

The common law system's foundation rests upon the principle of \textit{stare decisis}, mandating judicial adherence to precedents established in prior rulings when addressing analogous issues and facts within the same jurisdiction. As legal documentation grows in complexity and volume, sophisticated Natural Language Processing (NLP) techniques become indispensable for understanding, analyzing, and retrieving relevant precedents. \textbf{TraceRetriever} plays a crucial role in upholding \textit{stare decisis}, facilitating the identification of past judgments with similar legal contexts to ensure consistent application of the law. The sheer volume of legal resources, including judgments, statutes, and regulations, poses a significant challenge for legal professionals seeking pertinent precedents, underscoring the urgent need for effective retrieval mechanisms.

\begin{figure*}[t]
    \centering
    \includegraphics[width=\linewidth]{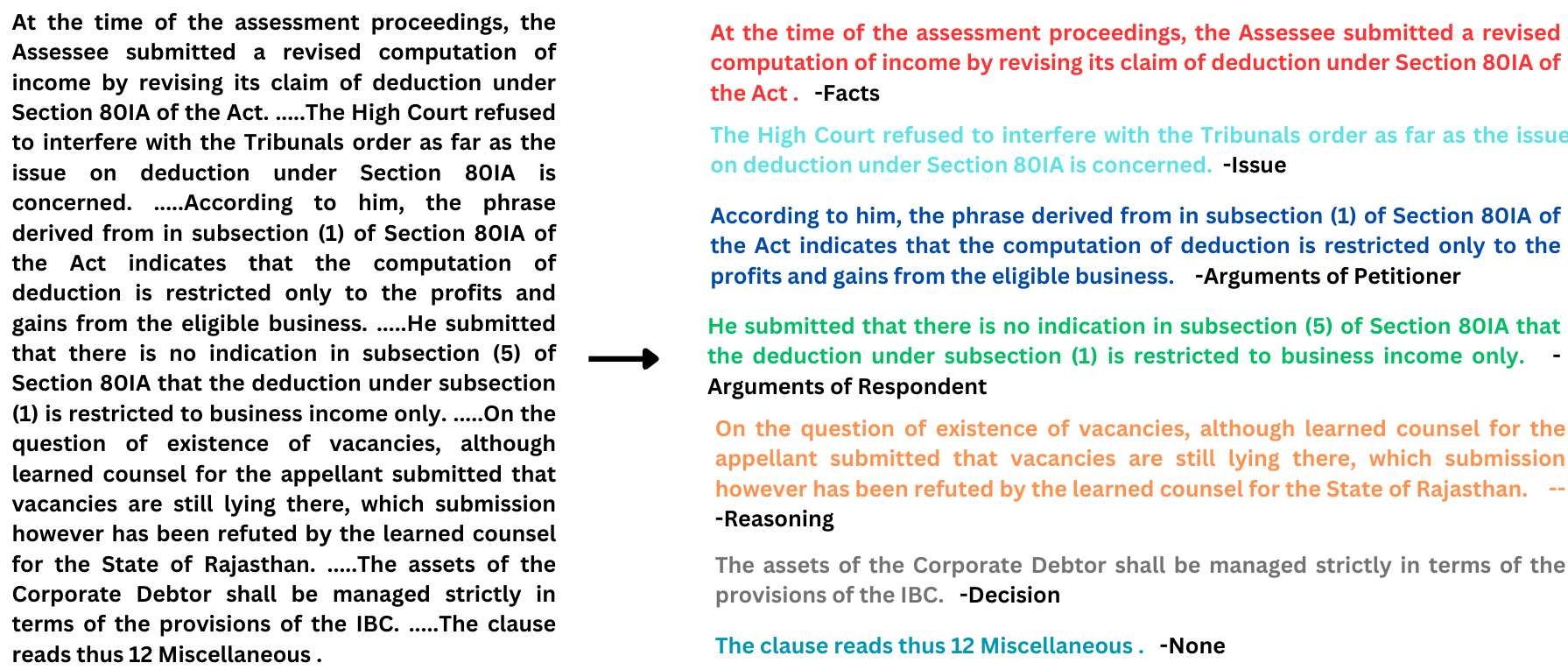}
    \caption{Illustration of rhetorical role segmentation in a legal document. The left side shows the original excerpt, while the right side displays the labeled segments. In our approach, only relevant segments such as \textit{Facts} and \textit{Issue} are retained to emulate real-world legal case retrieval scenarios, where complete information like \textit{Reasoning} or \textit{Decision} may not be available at query time \citep{nigam2025legalseg}.}
    \label{fig:RR_examples}
\end{figure*}

A notable limitation in much of the existing work on automated precedent retrieval is its reliance on using entire prior case documents as queries. This approach deviates significantly from real-world legal practice, where lawyers typically formulate search queries based on specific factual details and legal issues extracted from the case at hand, often with limited initial information. To address this gap, this paper tackles the challenge of mimicking real-world legal search scenarios in TraceRetriever by proposing a novel heuristic approach. Our methodology strategically integrates the complementary strengths of a keyword-based model (BM25), a semantic Vector Database, and a fine-grained Cross-Encoder for re-ranking. A key innovation of our work lies in utilizing a trained Hierarchical Bidirectional LSTM (HierBiLSTM) model by \citep{bhattacharya2019identification} to classify sentences within legal documents into distinct rhetorical roles. We then leverage the role segments, identified through this classification, as the query for our retrieval pipeline. This deliberate use of limited, rhetorically-informed query components directly mirrors the information scarcity often encountered in practical legal research. The core problem this paper addresses is therefore the development of a TraceRetriever system that effectively operates with limited, contextually relevant information, thereby more accurately reflecting real-world legal search processes.

To evaluate the effectiveness of our proposed TraceRetriever pipeline, we conducted experiments on two established legal datasets: the Indian Legal Text Understanding and Reasoning (IL-PCR) dataset \citep{joshi-etal-2023-u} and the Competition on Legal Information Extraction and Entailment (COLIEE) 2025 dataset. Our pipeline employs a heuristic approach that strategically integrates the strengths of three distinct retrieval models: a semantic Vector Database, the BM25 algorithm, and a more nuanced Cross-Encoder. To further refine the initial retrieval results from the Vector Database and BM25, we implemented Reciprocal Rank Fusion (RRF), a robust re-ranking technique.

In our TraceRetriever pipeline, we established BM25 as a robust baseline, representing a traditional keyword-based approach to information retrieval. To enhance the relevance and accuracy of our results, we implemented a sophisticated re-ranking strategy that leverages both semantic understanding and fine-grained interaction. Specifically, we employed Cross-Encoders to re-rank the top-k documents initially retrieved by two distinct methods: the lexical matching of BM25 and the semantic similarity captured by our Vector Database (a bi-encoder-based approach). This multi-faceted strategy effectively integrates the strengths of three complementary retrieval paradigms:

Our key contributions are:
\begin{enumerate}
\item A realistic legal retrieval strategy using rhetorical role-based queries reflecting limited-information scenarios.
\item Development of TraceRetriever: A hybrid pipeline integrating BM25, vector search, and cross-encoder re-ranking.
\end{enumerate}

For the sake of reproducibility, we have made our dataset, code, and RAG-based pipeline implementation via an github repository\footnote{\href{https://github.com/ShubhamKumarNigam/Legal_IR}{\texttt{https://github.com/ShubhamKumarNigam/Legal\_IR}}}. 

\section{Related Work}

Legal case retrieval has witnessed a rapid transformation with the advent of LLMs, RAG pipelines, and rhetorical role labeling. Traditionally, legal information retrieval relied heavily on lexical matching (e.g., BM25), which struggled to handle the semantic and structural nuances of legal texts. Recent innovations focus on improving retrieval accuracy by leveraging domain-specific embeddings, legal document structures, and rhetorical role understanding.

Several systems have explored enhancing legal QA and retrieval using hybrid architectures. \citep{10.1007/978-3-031-63646-2_29} integrates Case-Based Reasoning with RAG to improve contextual relevance and factual correctness in legal question-answering. Similarly, \citep{panchal2025lawpalretrievalaugmented} utilizes FAISS and DeepSeek embeddings to make Indian legal knowledge accessible through a chatbot interface.

Another significant trend is the use of rhetorical roles in structuring legal texts. \citep{bhattacharya2019identification, malik-etal-2022-semantic} pioneered rhetorical role classification in Indian legal judgments, showing that deep neural architectures such as BiLSTM-CRF and multi-task learning can outperform traditional methods. \citep{marino2023automatic} further advanced this by stacking transformers over LEGAL-BERT to capture inter-sentence dependencies for rhetorical role classification across multilingual legal datasets. These works collectively demonstrate the feasibility and utility of segmenting legal documents into roles such as \textit{Facts}, \textit{Issues}, and \textit{Reasoning} categories that are highly valuable for information extraction and retrieval. In recent studies, \citep{bhattacharya2019identification} proposed a CRF-BiLSTM model specifically for as
signing rhetorical roles to sentences in Indian legal
documents.

In the context of document-to-document legal retrieval, methods like \citep{althammer2022parmparagraphaggregationretrieval}, \citep{ma2023caseencoderknowledgeenhancedpretrainedmodel}, and \citep{li2023sailerstructureawarepretrainedlanguage} aim to overcome the challenges of long input lengths and weak semantic relevance by employing paragraph aggregation, structure-aware pretraining, and custom contrastive loss functions. Meanwhile, \citep{tang2023casegnngraphneuralnetworks} and \citep{tang2024caselinkinductivegraphlearning} take a graph based approach, modeling the connectivity between cases via attributed case graphs or global semantic networks to achieve state-of-the-art performance.
\citep{nigam2022nigam} presents a cascaded retrieval framework that integrates BM25 for lexical matching with Sentence BERT and Sent2Vec for semantic understanding. Interestingly, results show that BM25 alone often outperforms neural models, reaffirming the robustness and relevance of lexical approaches in legal case retrieval.

Beyond traditional lexical and semantic methods, several recent studies have explored innovative architectures to enhance legal case retrieval by addressing challenges such as long document length, complex legal semantics, and noisy or sparse queries. \citep{https://doi.org/10.1155/2022/2511147} proposes a retrieval method grounded in legal facts by combining topic modeling with BERT-based paragraph aggregation, offering more accurate semantic representations tailored to the legal domain. Similarly, \citep{ijcai2020p484} focuses on paragraph-level interactions, modeling fine-grained relationships between query and candidate cases to improve relevance estimation using a cascade framework and BERT finetuned on legal entailment tasks. Addressing structural and causal reasoning, \citep{10.1145/3580305.3599273} introduces a counterfactual graph learning approach, which transforms legal cases into graphs of legal elements and enhances retrieval via counterfactual data augmentation and relational graph neural networks. Meanwhile, \citep{zhou2023boostinglegalcaseretrieval} employ large language models (LLMs) to distill salient query content, showing that query reformulation using LLMs improves retrieval even in long, noisy legal queries. Structural reasoning is also emphasized in SLR \citep{zhou2023boostinglegalcaseretrieval}, which incorporates both internal (document segmentation into roles like Facts, Holding, Decision) and external (charge relationship graphs) structures to enhance retrieval accuracy via a learning-to-rank approach, \citep{santosh2025lecopcrlegalconceptguidedprior} enhances prior case retrieval by generating legal concepts from the factual section of a query case to capture semantic intent. 
Collectively, these works highlight a growing trend toward structurally aware, semantically enriched, and role-sensitive retrieval models supporting the need for rhetorical role-driven query formulations in real-world legal search settings.

While these systems improve retrieval through structure, semantics, or scale, few explicitly address the \textit{limited-information retrieval scenario} commonly encountered in real-world legal practice, where queries often arise from partial knowledge, such as only the \textit{Facts} or \textit{Issues} of a case. The \citep{deng-etal-2024-learning} framework approaches this partially by reformulating legal documents into interpretable sub-facts using LLMs, but it does not explicitly tie these sub-facts to rhetorical roles.

In contrast to general-purpose document retrieval, \citep{joshi-etal-2023-u} propose U-CREAT, an unsupervised retrieval framework that extracts and matches event tuples consisting of predicates and their arguments from entire legal documents. However, U-CREAT still requires parsing the full document to extract events and does not leverage explicit legal segmentation such as rhetorical roles.


\section{Task Description}
\label{sec:task_description}
The goal of this task is to develop models capable of retrieving the most relevant prior legal cases for a given query case, with a novel emphasis on mimicking realistic legal reasoning workflows. Unlike previous work that provides entire case documents as input queries to retrieval models, we constrain the query representation by leveraging rhetorical role segmentation. This segmentation reflects how legal professionals typically reason over and search with focused portions of a case, such as facts, issues, or arguments, rather than the full text.

Let $Q = \{q_1, q_2, \dots, q_p\}$ be a set of query legal cases, where each $q_i$ is a segmented case document composed of rhetorical roles:

\[
q_i = \{\texttt{Facts}_i, \texttt{Issues}_i, \texttt{Arguments}_i, \dots\}
\]

Rather than passing the full $q_i$ as a monolithic document, we present the segmented roles (individually or in combination) to retrieval models to enable fine-grained relevance modeling. This design encourages the system to focus on legally salient information while ignoring irrelevant or verbose content, thus improving efficiency and interpretability.

Let $D = \{d_1, d_2, \dots, d_n\}$ be a corpus of precedent legal documents. The objective is to retrieve a ranked list of $k$ relevant documents $R_i = \{r_{i1}, r_{i2}, \dots, r_{ik}\} \subseteq D$ for each query $q_i$, where documents are ranked by their relevance.

We define a retrieval scoring function:

\[
g: Q \times D \rightarrow \mathbb{R}
\]

where $g(q_i, d_j)$ outputs a relevance score indicating the degree to which the prior legal document $d_j$ is relevant to the query $q_i$. The retrieved list $R_i$ for a query $q_i$ is then constructed by selecting the top $k$ documents from $D$ based on their relevance scores:
\[
R_i = \text{top-}k \{d_j \in D \mid g(q_i, d_j) \text{ is high}\}
\]
The input to the system is a legal query $q_i$, and the output is a ranked list of $k$ prior legal documents $R_i$, ordered by their relevance to the query.



\section{Dataset}
To support research in the domain of Prior Case Retrieval (PCR), we utilize the IL-PCR (Indian Legal Prior Case Retrieval) corpus, a large-scale collection of Indian legal documents comprising 7,070 English-language case texts by \citep{joshi-etal-2023-u}. This corpus enables the development and benchmarking of retrieval systems specifically tailored to the Indian legal system.

\begin{table}[h]
\centering
\resizebox{\linewidth}{!}{%
\begin{tabular}{@{}lrr@{}}
\toprule
\textbf{Dataset} & \textbf{COLIEE’25} & \textbf{IL-PCR} \\
\midrule
\# Documents & 9498 & 7070 \\
Avg. Document Size & 4759.79 & 8093.19 \\
\# Query Documents & 2077 & 1182 \\
Vocabulary Size & 426,118 & 113,340 \\
Total Citation Links & 8640 & 8008 \\
Avg. Citations per Query & 4.16 & 6.775 \\
Language & English & English \\
Legal System & Canadian & Indian \\
\bottomrule
\end{tabular}
}
\caption{Comparison of the IL-PCR corpus \citep{joshi-etal-2023-u} with the COLIEE'25 dataset.}
\label{tab:dataset-comparison}
\end{table}

\subsection{Overview of Dataset}

The IL-PCR corpus was created by collecting case documents from the public domain through the IndianKanoon website\footnote{\url{https://indiankanoon.org/}}. The initial set comprises the 100 most-cited Supreme Court of India (SCI) judgments, referred to as the \textit{zero-hop set}. To increase citation density, cases cited within these judgments (the \textit{one-hop set}) were also collected. This hierarchical collection approach ensures that each document has multiple cited cases, allowing for robust retrieval evaluation \citep{joshi-etal-2023-u}. Following standard preprocessing, empty or invalid cases were discarded. The resulting corpus was partitioned into training (70\%), validation (10\%), and test (20\%) splits.

\subsection{Preprocessing}

The preprocessing pipeline includes named entity normalization using spaCy's NER model, alongside a manually curated gazetteer. This standardization improves the generalizability of learned representations. Hyperlinked citations in the documents were replaced with a standardized token \texttt{<CITATION>}, while references to statutes and laws were retained, aligning with the task focus on case retrieval rather than statute retrieval. Additionally, an alternate version of the dataset removes entire sentences containing citations, as discussed in \citep{joshi-etal-2023-u}.

\section{Methodology}

This section elucidates the TraceRetriever methodology, a multi-stage framework designed for effective prior case retrieval, particularly when initiated with partial case details. Our approach integrates advanced NLP techniques, starting with rhetorical role annotation to enable targeted querying of key document sections. We then employ a hybrid retrieval strategy, combining semantic vector search with lexical BM25 matching on a focused candidate set. The resulting ranked lists are fused using RRF, followed by a deep semantic re-ranking via a cross-encoder.

\subsection{Rhetorical Role Annotation of Legal Documents}
The initial stage of our methodology involves enriching legal documents with rhetorical role annotations at the sentence level. To achieve this, we first perform sentence segmentation using the spaCy library. We implement the BiLSTM-CRF architecture introduced by \citep{bhattacharya2019identification}, which integrates a BiLSTM network with a Conditional Random Field (CRF) layer. The model takes as input sentence embeddings generated using a sent2vec model trained specifically on Indian Supreme Court judgments. These embeddings are processed by the BiLSTM to capture the sequential context across sentences. The CRF layer then models the dependencies between adjacent labels, enabling the output to follow the inherent structural patterns present in legal documents. By leveraging contextual cues from surrounding sentences, the model assigns a rhetorical role label to each sentence in a coherent and structured manner. The output of this stage is a corpus of legal documents where each sentence is associated with a predicted rhetorical role, forming the foundation for subsequent information retrieval experiments.

\subsection{Vector Database Construction and Candidate Retrieval}

To enable efficient semantic retrieval of legal documents, we employed Milvus to store and query dense vector representations. Each entry in the collection comprised a unique \texttt{id}, a 768-dimensional embedding generated using the \href{https://huggingface.co/Snowflake/snowflake-arctic-embed-m-v2.0}{Snowflake Arctic Embed v2.0} model, and the original document text (limited to 60,000 characters). An IVF-FLAT index, configured with nlist = 2048 and using L2 distance, was built to facilitate rapid approximate nearest neighbor search. Query vectors, embedded using the same model, were matched against the collection, with the nprobe parameter controlling the search depth across partitions. The top-$k$ semantically similar documents were retrieved based on L2 distance, forming the candidate set for downstream re-ranking via cross-encoders. This stage ensures that initial retrieval captures documents with high semantic alignment to the input query.

\begin{figure*}[t]
    \centering
    \includegraphics[width=\linewidth]{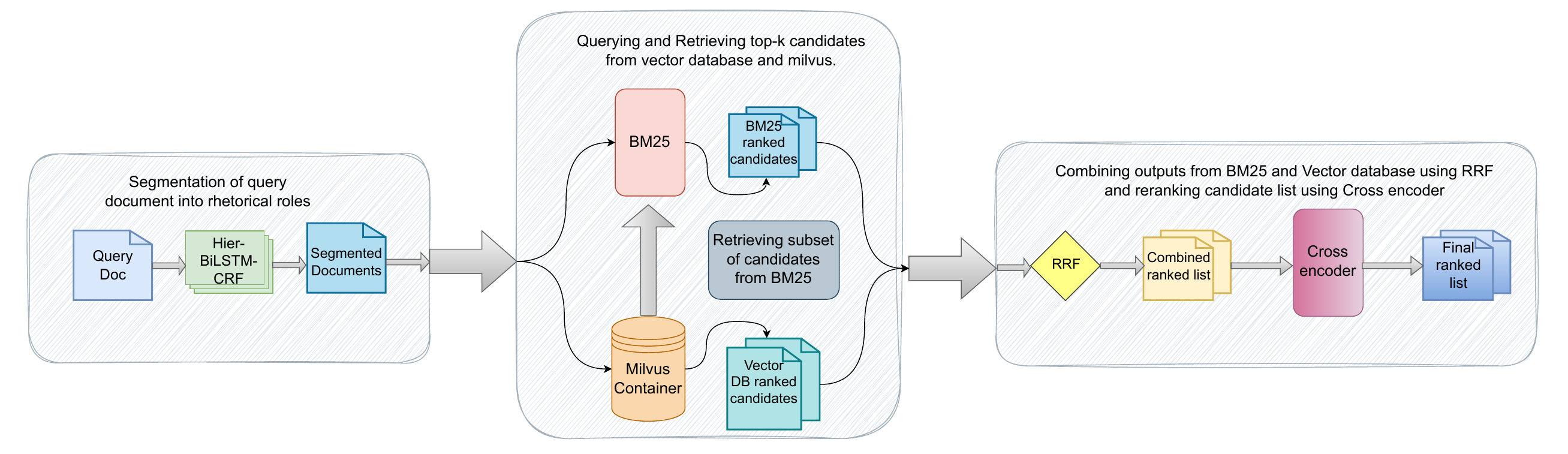}
    \caption{TraceRetriever Pipeline}
    \label{fig:Traceretriever}
\end{figure*}

\subsection{BM25 Retrieval on Vector Database Candidates}

To complement semantic similarity with lexical matching, BM25 is applied but only to a reduced candidate set to avoid high computational costs. These candidates are pre-selected using vector-based retrieval, ensuring that BM25 is run only on semantically relevant documents, balancing efficiency and retrieval accuracy. The process begins by selecting the top-$k$ candidates from the vector search. The parameter $k$ controls the trade-off between recall and efficiency larger $k$ may improve recall but increases computational load. We selected $k$ as 1000 to maintain this balance. BM25 then scores each candidate based on term frequency (TF) and inverse document frequency (IDF), ranking documents where rare, frequent query terms appear. This yields a refined list of documents ranked by lexical relevance. By applying BM25 only to vector-selected candidates, the system enhances semantic matching with precise lexical signals.

\subsection{Reciprocal Rank Fusion (RRF)}

To combine the ranked outputs from vector-based and BM25 retrieval, we employ Reciprocal Rank Fusion (RRF), a rank aggregation technique that leverages the complementary strengths of different retrieval methods for improved performance. Each document in the ranked lists receives a numerical rank (1 for top, 2 for second, etc.). Its reciprocal rank is computed as $\frac{1}{\text{rank} + k}$, where $k$ is a constant used to reduce the influence of lower-ranked results. We selected an optimal $k$ to balance influence across both retrieval methods. For each document, its reciprocal ranks across all lists are summed to generate an aggregated RRF score. Documents are then sorted in descending order of this score, producing a fused ranking that integrates both semantic similarity (from the vector DB) and lexical relevance (from BM25). RRF enhances retrieval by combining diverse signals, resulting in a more robust and accurate final document ranking than either method alone.

\subsection{Cross-Encoder Re-ranking}

To refine the ranking of candidate documents and prioritize the most relevant prior cases, we use a cross-encoder model. Unlike bi-encoders used in the initial retrieval, cross-encoders attend to both the query and document simultaneously. The process begins by forming (query, document) pairs from the top results obtained via Reciprocal Rank Fusion (RRF). This narrows the focus to promising candidates. Each pair is scored using the pre-trained \href{https://huggingface.co/BAAI/bge-reranker-v2-m3}{bge-reranker-v2-m3} model, which excels at capturing fine-grained semantic interactions. For long documents exceeding the model’s input limits, a chunking strategy is applied. Each chunk is scored individually, and a final relevance score is computed using a weighted average of chunk scores. Other aggregation strategies like max or mean can also be used. Finally, documents are re-ranked based on these cross-encoder scores. This yields a final ranked list where the most semantically relevant cases are prioritized, enhancing retrieval quality by leveraging the model’s deep understanding of query-document relations.

\subsection{TraceRetriever: A Hybrid Legal Case Retrieval Framework}

The TraceRetriever pipeline combines rhetorical role segmentation, vector-based retrieval, keyword-based retrieval (BM25), reciprocal rank fusion (RRF), and cross-encoders to perform effective and realistic legal case retrieval. It begins by segmenting the query legal document into sentences and classifying each into rhetorical roles (e.g., \textit{Facts}, \textit{Issue}, \textit{Argument}, \textit{Reasoning}, and \textit{Decision}) using a pre-trained Hierarchical BiLSTM. This segmentation supports role-specific querying, reflecting real-world scenarios where legal practitioners often search based on partial case descriptions. To retrieve initial candidates efficiently, a bi-encoder is used to encode both the rhetorically-filtered query and documents into dense embeddings. A vector database is then queried to retrieve the top-$k$ semantically relevant documents. Since applying BM25 across the entire corpus is computationally expensive, it is selectively applied only to this subset of vector-retrieved documents to capture lexical overlap. To unify the strengths of semantic and lexical signals, the results from the vector search and BM25 are merged using Reciprocal Rank Fusion (RRF), which produces a single ranked list. Finally, a cross-encoder re-ranks this list by jointly encoding each query-document pair to compute fine-grained relevance scores. Through this multi-stage approach, TraceRetriever effectively combines semantic understanding, lexical precision, and deep relevance modeling addressing the challenges of prior case retrieval under limited-information conditions.

\section{Evaluation Metrics}

To evaluate the effectiveness of our information retrieval models, we employ a standard set of metrics commonly used in retrieval tasks.

Our primary evaluation relies on Precision@k, Recall@k, Mean Average Precision (MAP), Mean Reciprocal Rank (MRR), and F1@k. Precision@k quantifies the fraction of relevant documents within the top-k retrieved results, whereas Recall@k assesses the system's capability to identify all relevant documents within the top-k. MAP offers an overall performance measure by averaging the precision at each rank where a relevant document is found, across all queries. MRR focuses on the rank of the first relevant document in the result list. Finally, F1@k calculates the harmonic mean of Precision@k and Recall@k, providing a balanced evaluation of both aspects. Collectively, these metrics offer a thorough evaluation framework for assessing the ranking effectiveness and retrieval performance of the models.
Here, we introduce the results of our experiments and discuss the performance of various models. Table \ref{TraceRetreiver} provides a summary of evaluation metrics for every model.

\section{Results Analysis}

Our experimental evaluation demonstrates significant variations in retrieval performance across different query formulations based on rhetorical roles and retrieval methodologies. Table~\ref{TraceRetreiver} presents a comprehensive comparison of precision, recall, F1-score, Mean Average Precision (MAP), and Mean Reciprocal Rank (MRR) across all experimental configurations.

\subsection{Retrieval Method Performance}
The empirical results reveal distinct performance characteristics among the three retrieval methods. BM25, a traditional lexical matching approach, consistently underperforms compared to the semantic-based methods across all query configurations. This performance gap underscores the limitations of term-frequency based approaches in capturing the nuanced legal semantics present in case documents. Vector DB demonstrates superior performance in precision-oriented metrics, achieving the highest MAP (0.3783) and MRR (0.3924) scores with the \textit{Facts+Issue+Reasoning} configuration. Notably, Vector DB consistently requires lower optimal $k$ values (typically 5--7), indicating its strong ability to position relevant documents at higher ranks. This characteristic makes Vector DB particularly suitable for applications where precision at lower ranks is prioritized. The Cross-encoder model exhibits different performance characteristics, consistently achieving higher recall values but requiring larger $k$ values (7--11) to reach optimal performance. For instance, with the \textit{Facts+Issue+Reasoning} configuration, the Cross-encoder achieves the highest recall (0.2815) among all methods but at $k = 11$. This suggests that Cross-encoder captures a broader range of relevant documents but with less precise ranking capability compared to Vector DB.

\begin{table}[t]
\centering
\resizebox{\linewidth}{!}{%
\begin{tabular}{l@{\hskip 1pt}l@{\hskip 1pt}c@{\hskip 4pt}c@{\hskip 4pt}c c c c}
\hline
\toprule
\textbf{Dataset} & \textbf{Model} & \textbf{Precision@k} & \textbf{Recall@k} & \textbf{F1-score@k} & \textbf{MAP} & \textbf{MRR} & \textbf{$k$} \\
\hline
\\
\multirow{3}{*}{\shortstack{Full Query\\(IL-PCR)}}
& BM25          & 0.0819 & 0.1023 & 0.0740 & 0.2116 & 0.2182 & 6\\
& Vector DB     & 0.1715 & 0.1754 & 0.1419 & 0.3484 & 0.3585 & 5\\
& Cross-encoder & 0.1459 & 0.1858 & 0.1301 & 0.3480 & 0.3339 & 6\\\\
\hline
\\
\multirow{3}{*}{\shortstack{Facts\\(IL-PCR)}}
& BM25 & 0.0797 & 0.0835 & 0.0694 & 0.1599 & 0.1684 & 5\\
& Vector DB & 0.1093 & 0.1574 & 0.1097 & 0.2566 & 0.2783 & 7\\
& Cross-encoder & 0.0916 & 0.2050 & 0.1082 & 0.2364 & 0.2725 & 11\\\\
\hline
\\
\multirow{3}{*}{\shortstack{Facts+\\Issue\\(IL-PCR)}}
& BM25 & 0.0803 & 0.1152 & 0.0800 & 0.1907 & 0.2014 & 7\\
& Vector DB & 0.1281 & 0.1606 & 0.1200 & 0.2880 & 0.3055 & 6\\
& Cross-encoder & 0.1134 & 0.1723 & 0.1143 & 0.2554 & 0.2733 & 7\\\\
\hline
\\
\multirow{3}{*}{\shortstack{Facts+\\ Issue+\\Arguments\\(IL-PCR)}}
& BM25 & 0.0900 & 0.1328 & 0.0908 & 0.2111 & 0.2259 & 7\\
& Vector DB & 0.1630 & 0.1775 & 0.1418 & 0.3291 & 0.3431 & 5\\
& Cross-encoder & 0.1121 & 0.2295 & 0.1277 & 0.2680 & 0.3045 & 10\\\\
\hline
\\
\multirow{3}{*}{\shortstack{Facts+\\Issue+\\Reasoning\\(IL-PCR)}}
& BM25 & 0.0947 & 0.1034 & 0.0824 & 0.2081 & 0.2144 & 5\\
& \textbf{Vector DB} & \textbf{0.1843} & \textbf{0.2088} & \textbf{0.1636} & \textbf{0.3783} & \textbf{0.3924} & \textbf{5}\\
& Cross-encoder & 0.1223 & 0.2815 & 0.1436 & 0.2973 & 0.3316 & 11\\\\
\hline
\\
\multirow{3}{*}{\shortstack{Facts+\\Issue+\\Decision\\(IL-PCR)}}
& BM25 & 0.0884 & 0.1115 & 0.0833 & 0.1864 & 0.1926 & 6\\
& Vector DB & 0.121 & 0.1747 & 0.1212 & 0.2931 & 0.3157 & 7\\
& Cross-encoder & 0.1006 & 0.2235 & 0.1179 & 0.265 & 0.2991 & 11\\\\
\hline
\\

\multirow{3}{*}{\shortstack{Coliee\\Dataset}}
& BM25 & 0.0549 & 0.1139 & 0.0661 & 0.1410 & 0.1440 & 6\\
& Vector DB & 0.0515 & 0.1795 & 0.0720 & 0.1695 & 0.1786 & 11\\
& Cross-encoder & 0.0587 & 0.1545 & 0.0754 & 0.1574 & 0.1638 & 8\\\\
\hline
\end{tabular}}
\caption{Performance comparison across different query configurations and models on IL-PCR and COLIEE datasets}
\label{TraceRetreiver}
\end{table}

\subsection{Impact of Rhetorical Role Configurations}
The experimental results demonstrate that query formulation using specific rhetorical roles significantly impacts retrieval effectiveness. Several key observations emerge:

Using only factual components (\textit{Facts}) yields the lowest performance across all retrieval methods, with Vector DB achieving MAP of 0.2566 and MRR of 0.2783. This finding suggests that factual information alone provides insufficient context for effective legal case retrieval. The addition of issue information (\textit{Facts+Issue}) produces modest improvements across all models, with Vector DB showing MAP of 0.2880 and MRR of 0.3055. This improvement indicates that legal issues provide important discriminative information beyond mere facts. When argumentative elements are incorporated (\textit{Facts+Issue+Arguments}), we observe substantial performance gains, particularly for Vector DB (MAP: 0.3291, MRR: 0.3431) and Cross-encoder (Recall@k: 0.2295). This suggests that arguments contain substantive information about legal reasoning that aids in identifying relevant precedents. The \textit{Facts+Issue+Reasoning} configuration consistently yields the best performance across all retrieval methods, with Vector DB achieving the highest overall MAP (0.3783) and MRR (0.3924). This finding highlights the critical importance of legal reasoning components in determining case relevance. It suggests that the explicit reasoning articulated by judges forms the most discriminative aspect of legal documents for retrieval purposes. Interestingly, incorporating the decision component (\textit{Facts+Issue+Decision}) results in performance degradation compared to the reasoning configuration. Vector DB's MAP decreases to 0.2931 and MRR to 0.3157, while Cross-encoder shows similar declines. This degradation may be attributed to the fact that decisions often contain standardized language that is less discriminative than the specific reasoning that led to those decisions. The full query configuration performs relatively well (Vector DB: MAP 0.3484, MRR 0.3585), but still falls short of the \textit{Facts+Issue+Reasoning} configuration. This indicates that using the entire document introduces noise that dilutes retrieval effectiveness.

\subsection{Dataset Comparison}
A comparison between the IL-PCR and COLIEE datasets reveals substantial performance disparities. All retrieval methods perform markedly better on the IL-PCR dataset. On the COLIEE dataset, the best performance is achieved by Vector DB with MAP of 0.1695 and MRR of 0.1786, substantially lower than the corresponding metrics on IL-PCR. This disparity may be attributed to differences in document structure, domain-specific language, or the inherent complexity of the legal relationships represented in the COLIEE dataset. Additionally, our BiLSTM-based rhetorical role segmentation model was trained specifically on Indian legal documents.

\subsection{Optimal \texorpdfstring{$k$}{k} Values}
In the context of information retrieval, $k$ represents the number of top-ranked documents retrieved by a system. An interesting observation from our experiments is the variation in optimal $k$ values across different configurations. Vector DB generally achieves optimal performance at lower $k$ values (5--7), while Cross-encoder typically requires higher $k$ values (7--11) to reach optimal performance. This pattern is consistent across query configurations and further emphasizes the distinct characteristics of these retrieval approaches: Vector DB excels at precise ranking of highly relevant documents within a smaller top-$k$ set, while Cross-encoder captures a broader range of potentially relevant documents, often requiring a larger top-$k$ to include the most pertinent results due to less precise initial ranking.

\subsection{Error Analysis}

Retrieval errors were common when queries lacked argumentative depth or rhetorical coherence. Partial segments like \textit{Facts} or \textit{Facts+Issue} often led to vague queries, reducing the ability to retrieve precise legal precedents. Cross-encoders achieved high recall but lower MAP in such settings. For example, in the \textit{Facts-only} configuration (Table \ref{TraceRetreiver}), recall was 0.205, but MAP dropped to 0.2364, indicating difficulty in ranking the most legally relevant documents.

BM25 struggled with rhetorical overlap, particularly in IL-PCR, where \textit{Facts-only} and \textit{Facts+Issue} yielded low MAPs of 0.1599 and 0.1907. Its reliance on surface-level term frequency limited its ability to distinguish semantically similar yet legally distinct content. Interestingly, dense retrieval with Vector DB performed better in focused configurations. In IL-PCR, the MAP improved from 0.3484 (\textit{Full}) to 0.3783 (\textit{Facts+Issue+Reasoning}), likely due to reduced procedural noise and improved signal-to-noise ratio in embeddings. This suggests that full-document queries, though comprehensive, may dilute dense models with irrelevant content. In contrast, selected rhetorical segments enhance semantic richness and focus. Cross-encoders performed best when queries included \textit{Arguments} or \textit{Reasoning}, but struggled without structured argumentative flow. Overall, Vector DB benefited most from rhetorically rich inputs, with combinations like \textit{Facts+Issue+Reasoning} offering the best trade-off between semantic depth and legal specificity.

In COLIEE, absence of rhetorical segmentation degraded performance across models. Vector DB's MAP dropped to 0.1695, and BM25 to 0.141, as noisy, unsegmented queries confused both dense and sparse retrievers.
The rhetorical classifier, trained on Indian cases, also failed to generalize to Canadian judgments in COLIEE, reducing the effectiveness of rhetorical-aware retrieval.

\section{Conclusions and Future Work}

This work introduced a novel approach to prior case retrieval that better reflects real-world legal research, where professionals often rely on partial case information like \textit{Facts} and \textit{Issue}. By using rhetorical role segmentation to extract these components as queries, our method simulates realistic legal workflows. Evaluations on ILTUR and COLIEE datasets showed that even under these constraints, our pipeline BM25, VectorDB, RRF, and cross-encoder reranking retrieves relevant cases, though with reduced precision and recall compared to full-document queries. Nonetheless, this role-based querying aligns closely with how legal professionals conduct research, offering a practical shift in retrieval methodology. Our main contribution is a conceptual framework for retrieval under partial information, encouraging a more practice-oriented direction in legal IR. Rather than chasing ideal scores, we aim to model realistic scenarios that support practical system design. This work has laid the groundwork for a more realistic paradigm in prior case retrieval by focusing on the information actually available at the initial stages of legal research. Our findings underscore the viability of a pipeline leveraging rhetorical role segmentation for query formulation, demonstrating effective, albeit reduced, retrieval performance compared to methods relying on complete case documents. Future work includes improving retrieval robustness under sparse queries, enhancing rhetorical segmentation, and testing advanced rerankers. We also aim to explore cross-lingual and multi-domain retrieval to further bridge academic research and real-world legal use cases.

\section*{Limitations}

While this work presents a novel approach to prior case retrieval that mirrors real-world legal research, several limitations remain and highlight directions for improvement. A key challenge is the semantic sparsity of queries constructed from only rhetorical roles like \textit{Facts} and \textit{Issue}. This constrained input can omit important context, limiting the models’ ability to fully capture legal reasoning and reducing retrieval precision. Rhetorical overlap between roles such as \textit{Facts} and \textit{Reasoning} poses another issue. Their linguistic similarity makes it difficult especially for models like BM25 to differentiate cases based solely on rhetorical cues. While cross-encoders and vector models mitigate this to some extent, they still struggle with nuanced legal distinctions. Class imbalance in rhetorical roles also affects performance, particularly for underrepresented roles like \textit{Issue} or \textit{Decision}. Additionally, the computational complexity of advanced models like cross-encoders and dense retrievers can hinder scalability. Their high resource demands may limit deployment in real-world systems. Future work should explore optimization techniques such as pruning or quantization to maintain performance with lower resource requirements. While the system shows promise under real-world constraints, addressing these limitations will be crucial for building scalable and robust legal retrieval systems.

\newpage
\bibliography{sknigam}

\end{document}